\documentclass[
               final,
               twocolumn,
               11pt]{elsarticle_nonatbib}

\usepackage{amssymb}

\usepackage{lineno}

\usepackage[
  natbib = true,
    backend=bibtex,
    isbn=true,
    url=true,
    doi=true,
    eprint=false,
    style=numeric-comp,
    sorting=none,
    sortcites = true,
    maxcitenames=1,
    minbibnames = 3,
    maxbibnames = 6
           ]{biblatex}
\DeclareFieldFormat{labelnumberwidth}{#1\adddot}
\bibliography{bibliography}

\DeclareGraphicsExtensions{
    .png,.PNG,.jpg,.jpeg,.JPG,.JPEG,
    .pdf,.PDF,.eps,.EPS,.ps,.PS
  }

\journal{arXiv}

\usepackage{lipsum}
\usepackage{booktabs}
\usepackage{listings}
\usepackage{color}
\definecolor{listlightgray}{gray}{0.955}
\usepackage[style=altlist]{glossaries}
\makeglossaries

\usepackage{todonotes}
\usepackage{float}

\newacronym{CT}{CT}{computed tomography}
\newacronym{MR}{MR}{magnetic resonance}
\newacronym{MRI}{MRI}{magnetic resonance imaging}
\newacronym{ESV}{ESV}{end-systolic volume}
\newacronym{EDV}{EDV}{end-diastolic volume}
\newacronym{VM}{VM}{ventricular mass}
\newacronym{EF}{EF}{ejection fraction}
\newacronym{SV}{SV}{stroke volume}
\newacronym{LV}{LV}{left-ventricle}
\newacronym{RV}{RV}{right-ventricle}
\newacronym{NIH}{NIH}{National Institutes of Health}
\newacronym{CNMC}{CNMC}{Children's National Medical Center}
\newacronym{cv}{cv}{chamber view}
\newacronym{nifti}{NIFTI}{Neuroimaging Informatics Technology Initiative}
\newacronym{RMSD}{RMSD}{root mean square distance}
\newacronym{MAPE}{MAPE}{mean absolute percentage error}
\newacronym{dice}{DSC}{Dice similarity coefficient}
\newacronym{ME}{ME}{mean error}
\newacronym{MAPD}{MAPD}{mean absolute percentage error}
\newacronym{relu}{ReLU}{rectified linear unit}
\newacronym{prelu}{PReLU}{parametric rectified linear unit}
\newacronym{RVSC}{RVSC}{Right Ventricular Segmentation Challenge}
\newacronym{LVSC}{LVSC}{MICCAI 2009 LV Segmentation Challenge}
\newacronym{kaggle}{DSBCC}{Data Science Bowl Cardiac Challenge}
\newacronym{mhh}{MHH}{Hannover Medical School}
\newacronym{ICC}{ICC}{intraclass correlation coefficient}
\newacronym{GPU}{GPU}{graphics processing unit}
\newacronym{CPU}{CPU}{central processing unit}
\newacronym{OpenMP}{OpenMP}{Open Multiprocessing}
\newacronym{CUDA}{CUDA}{Compute Unified Device Architecture}
\newacronym{ssfp}{ssfp}{steady state free precession}

\begin{document}
\begin{frontmatter}

\title{$\nu$-net: Deep Learning for Generalized Biventricular Cardiac Mass and Function Parameters}

\auth[mhhrad,breath]{Hinrich B Winther\corref{cor1}\corref{cor2}}
\auth[jguinf]{Christian Hundt\corref{cor2}}
\auth[jguinf]{Bertil Schmidt}
\auth[mhhrad,breath]{Christoph Czerner}
\auth[mhhcar]{Johann Bauersachs}
\auth[mhhrad,breath]{Frank Wacker}
\auth[mhhrad,breath]{Jens Vogel-Claussen}

\address[mhhrad]{Institute for Diagnostic and Interventional Radiology, Hannover Medical School, Hannover, Germany}
\address[breath]{Biomedical Research in Endstage and Obstructive Lung Disease Hannover (BREATH), Member of the German Center for Lung Research, Hannover, Germany.}
\address[jguinf]{Institute for Computer Science, Johannes Gutenberg University, Mainz, Germany}
\address[mhhcar]{Department of Cardiology and Angiology, Hannover Medical School, Hannover, Germany \vspace{-1cm}}

\cortext[cor1]{hbwinther@metalabs.de}
\cortext[cor2]{authors contributed equally}

\begin{abstract}
\noindent \textbf{Background:}
Cardiac MRI derived biventricular mass and function parameters, such as end-systolic volume (ESV), end-diastolic volume (EDV), ejection fraction (EF), stroke volume (SV), and ventricular mass (VM) are clinically well established. Image segmentation can be challenging and time-consuming, due to the complex anatomy of the human heart.
 
\noindent \textbf{Objectives:}
This study introduces $\nu$-net (/nju:n$\varepsilon$t/) -- a deep learning approach allowing for fully-automated high quality segmentation of right (RV) and left ventricular (LV) endocardium and epicardium for extraction of cardiac function parameters.
 
\noindent \textbf{Methods:}
A set consisting of 253 manually segmented cases has been used to train a deep neural network. Subsequently, the network has been evaluated on 4 different multicenter data sets with a total of over 1000 cases.

\noindent \textbf{Results:}
For LV EF the intraclass correlation coefficient (ICC) is 98, 95, and 80~\% (95 \%), and for RV EF 96, and 87~\% (80~\%) on the respective data sets (human expert ICCs reported in parenthesis). The LV VM ICC is 95, and 94~\% (84~\%), and the RV VM ICC is 83, and 83~\% (54~\%). This study proposes a simple adjustment procedure, allowing for the adaptation to distinct segmentation philosophies. $\nu$-net exhibits state of-the-art performance in terms of dice coefficient.
 
\noindent \textbf{Conclusions:}
Biventricular mass and function parameters can be determined reliably in high quality by applying a deep neural network for cardiac MRI segmentation, especially in the anatomically complex right ventricle. Adaption to individual segmentation styles by applying a simple adjustment procedure is viable, allowing for the processing of novel data without time-consuming additional training.
\end{abstract}


\end{frontmatter}


\clearpage
\section{Introduction}
The World Health Organization identifies ischaemic heart diseases as the leading cause of death~\cite{who_top_ten_cod}. Imaging technologies, such as \gls{MRI}, yield clinically well established parameters, including \gls{ESV}, \gls{EDV}, \gls{EF} and \gls{SV} as well as \gls{VM}. In order to determine these clinical biventricular cardiac mass and function parameters, usually a skilled physician with expertise in cardiac \gls{MRI} has to segment the image data. This task is typically performed in a time span of approximately 15-20 minutes per case.

Automated image segmentation, especially of the 2D short axis cardiac cine \gls{MRI} stacks is a highly competitive research field. Active contour models~\cite{kass_snakes_1988, osher_fronts_1988}, and machine learning approaches~\cite{poudel_recurrent_2016, avendi_combined_2016, tran_fully_2016} are among the most successful methods.

One major challenge for the design of robust classifiers for automated cardiac image segmentation is the lack of manually annotated training data (ground truth). Hence, models with a high number of free parameters, such as deep neural networks, tend to overfit to the characteristics of the assembled data. Image segmentation quality for similar image morphologies is typically sufficient, however, the quality might rapidly degrade for differing image characteristics. This can be caused by varying image acquisition techniques, varying experimental protocols and image morphology altering illnesses, such as cardiomyopathies. Additionally, image segmentation philosophies have been shown to have major influence on the resulting biventricular mass and function parameters. The inclusion or exclusion of trabeculations and papillary muscles affect the left and right ventricular mass as well as the endocardial volume~\cite{sievers_impact_2004, winter_evaluating_2008, vogel-claussen_left_2006, kawel-boehm_normal_2015}. No convention has been universally accepted for analyzing trabeculation and papillary muscle mass~\cite{schulz-menger_standardized_2013}. Moreover, the amount of realizable medical images is orders-of-magnitude bigger than the number of assembled samples. Many studies limit themselves to a single data source at a time for training and validation~ \cite{avendi_combined_2016, queiros_fast_2014, hu_hybrid_2013, liu_automatic_2012, huang_image-based_2011}. This might impair the generalization potential of corresponding models. 

A recent paper by Tran et al.~\cite{tran_fully_2016} applied transfer learning to adapt a pre-trained model to novel data sets. Unfortunately, a major drawback of this approach is the time-consuming retraining of the neural network. 

Our study investigates the generalization potential of cardiac image segmentation. In detail, we have composed a diverse data set consisting of images with highly varying characteristics and further applied non-linear augmentation techniques to artificially increase the number of training samples by orders-of-magnitude. We further demonstrate how to determine fully automated high quality estimates of clinical parameters, such as end-systolic volume (ESV), end-diastolic volume (EDV), ejection fraction (EF), stroke volume (SV), and ventricular mass (VM).


\section{Materials and Methods}

\subsection{Evaluation Measures}
In this study we evaluate the performance of the proposed cardiac segmentation approach by determining the clinical gold standard parameters \gls{EF}, \gls{ESV}, \gls{EDV} as well as \gls{VM} of the left and right ventricle. Furthermore, we compare the automatically computed image segmentation with the ground truth in terms of similarity measures, such as overlap, and \gls{dice}.

\subsubsection{Evaluation of Biventricular Cardiac Mass and Function}
The performance of the computed segmentation is determined by calculating \gls{EF}, \gls{ESV} and \gls{EDV}. In order to determine the \gls{EF} in a 2D short axis cine \gls{MRI} stack, the \gls{ESV} and \gls{EDV} are typically measured by segmenting the corresponding volumes as described by \citet{winter_evaluating_2008} (Simpson’s method). The \gls{EF} is the fraction of the \gls{EDV}, which is ejected with every cardiac cycle, i.e. the normalized difference of the corresponding volumes at end-systole and end-diastole:

\begin{equation}
    \text{EF}[\%] = \frac{V^{ED}_{endo} - V^{ES}_{endo}}{V^{ED}_{endo}} \cdot 100
\label{eq:EF}
\end{equation}

\noindent where $V^{ES}_{endo}$ denotes the \gls{ESV} and $V^{ED}_{endo}$ refers to the \gls{EDV}.

The \acrfull{VM} is calculated as product of a constant conversion density factor $\rho$ and the volume of the right ($V^{RV}_{epi}$) or the left ($V^{LV}_{epi}$) ventricle, respectively:

\begin{equation}
    \text{VM} = \rho \cdot ( V_{epi} - V_{endo} )
\label{eq:VM}
\end{equation}

\noindent where $\rho$ is a phenomenologically determined conversion factor of 1.05~g/cm$^3$. Note that the \gls{VM} is usually determined at the end-diastole.

The clinical parameters are gathered for the manual ground truth and the automatically computed prediction by counting the corresponding voxels. Volumes and other derived quantities are compared in terms of Spearman's rank correlation coefficient, the \gls{RMSD}, \gls{MAPE}, \gls{ME} as well as \gls{ICC}.

\subsubsection{Evaluation of Technical Parameters}
In this study we present the expected performance metrics in order to evaluate the quality of the automated cardiac segmentation. These metrics include geometrical measures quantifying overlap based on the \gls{dice} and traditional Machine Learning metrics, such as accuracy, precision, recall, and specificity.

The \gls{dice} is proportional to the ratio of intersection between two volumes divided by the sum of said volumes:

\begin{equation}
    \text{DSC}(X,Y) = 2 \frac{|X \cap Y|}{|X| + |Y|}.
\label{eq:dice}
\end{equation}

All performance measures are carried out on calculated 3D spatial volumes.

\subsection{Data Sets}
The experiments were performed on four independent 2D short axis cine cardiac \gls{MRI} data sets as depicted in Figure~\ref{fig:split}:

\begin{figure*}[t]
    \centering
    \includegraphics[width=.6\linewidth]{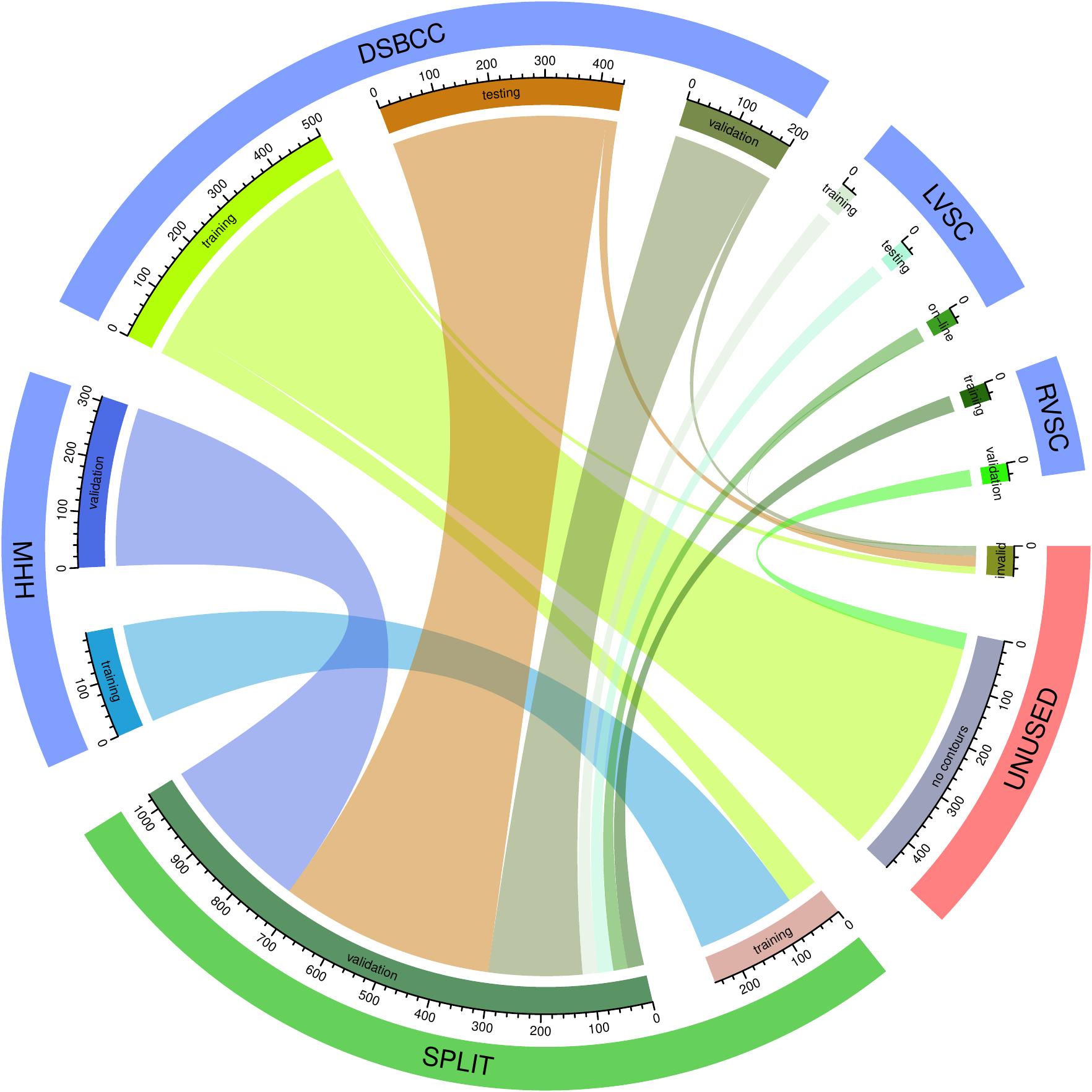}
    \caption{Depiction of the split used for training and validating the neural network as well as the corresponding data sources from \acrfull{mhh}, the \acrfull{kaggle}, the \acrfull{LVSC}, and the \acrfull{RVSC}. The outer blue sections denote the different data sources, the red section marks the portion of unused data, and the green section connotes the resulting training/validation split.}
    \label{fig:split}
\end{figure*}

\subsubsection{\gls{mhh} Data Set}
The data set consists of 193 training and 309 validation 2D \gls{ssfp} short axis cine \gls{MRI} stacks. The end-systolic and end-diastolic contours have been created by a senior radiologist (15 year experience in cardiac \gls{MRI}) and are accepted as ground truth for this study. Image acquisition was performed on three different 1.5~T \gls{MRI} scanners of a single vendor (Siemens Healthineers, Erlangen, Germany) using a \gls{ssfp} sequence with a slice thickness of 8~mm, no gap, an acquisition matrix of 256~$\times$~208 pixels with an in-plane spatial resolution of 1.4~$\times$~1.4~mm$^2$. 

\subsubsection{\gls{kaggle} Data}
The data set is composed of 500 training, 440 testing, and 200 validation 2D short axis cine stacks. It has been compiled by the \gls{NIH} and the \gls{CNMC}. It is at least one order-of-magnitude more extensive than any data set previously released. Each stack contains approximately 30 images over the cardiac cycle. The data set does not include a ground truth segmentation, instead, the end-systolic and end-diastolic volumes of the \gls{LV} are provided. This data set has been subject of the Second Annual Data Science Bowl contest.

Our study design does not require the fine-grained separation of the data into a test, training, and validation set as provided. Therefore, the validation and test set have been merged in order to extend the validation set to 640 cases. The 2D short axis cine \gls{MRI} stacks  have been converted into 4D matrices, stored in the \gls{nifti} format. This process failed for 14 of the training and 38 of the validation cases due to inconsistent image dimensions, resulting in 486 training and 602 validation \gls{nifti} files. 

The training set has been used twice. First, 60 cases with high visual diversity have been manually segmented and subsequently included in the training process of the neural network. Second, all 486 training cases have been utilized to fit a linear regression for the adjustment of the clinical parameters.

\subsubsection{\gls{LVSC} Data Set}
The MICCAI 2009 LV Segmentation Challenge~\cite{MICCAI_2009_challange_database} data set has been published by the Sunnybrook Health Sciences Center (Canada). This data set has been utilized in two separate experiments:

In the first experiment, this data set is used exclusively for validation purposes in this study, i.e. none of the images is used for training the neural network. This decision has been made in order to explore the generalization potential of the network and is being referred to as the ad-hoc performance. The data set consists of 45 cases. Ground truth segmentation is available for the end-diastole and the end-systole. It is composed of 12 heart failure with infarction, 12 heart failure without infarction, 12 \gls{LV} hypertrophy patients and 9 healthy subjects. 
The data was split into three parts: 15 for training, 15 for testing, and 15 for validation in an on-line contest. Data conversion has failed for one case (SC-HYP-37). As mentioned before, no training images of this data set have been used for this study. Therefore, all available images have been assembled into a single validation set consisting of 44 cases with end-systolic and end-diastolic segmentation, yielding a total of 88 2D image stacks.

In the second experiment, the data set has been split into 29 training and 15 validation images. A network, pretrained on the regime as depicted in Figure~\ref{fig:split}, has been retrained on this specific split, in order to evaluate the performance in contrast to the ad hoc results, and referred to as results with retraining.

Image acquisition was performed on a 1.5~T GE Signa \gls{MRI} scanner from the atrioventricular ring to the apex in 6 to 12 2D short axis cine stacks with a slice thickness of 8~mm, a gap of 8~mm, a field of view of 320~$\times$~320~mm$^2$ with a matrix resolution of 256~$\times$~256.

This data set is relevant since it has been extensively used in prior studies for training and evaluation. The end-systolic and end-diastolic contours are provided for the endo- and epicardial volume.

\begin{figure*}[t]
    \centering
    \includegraphics[width=.8\linewidth]{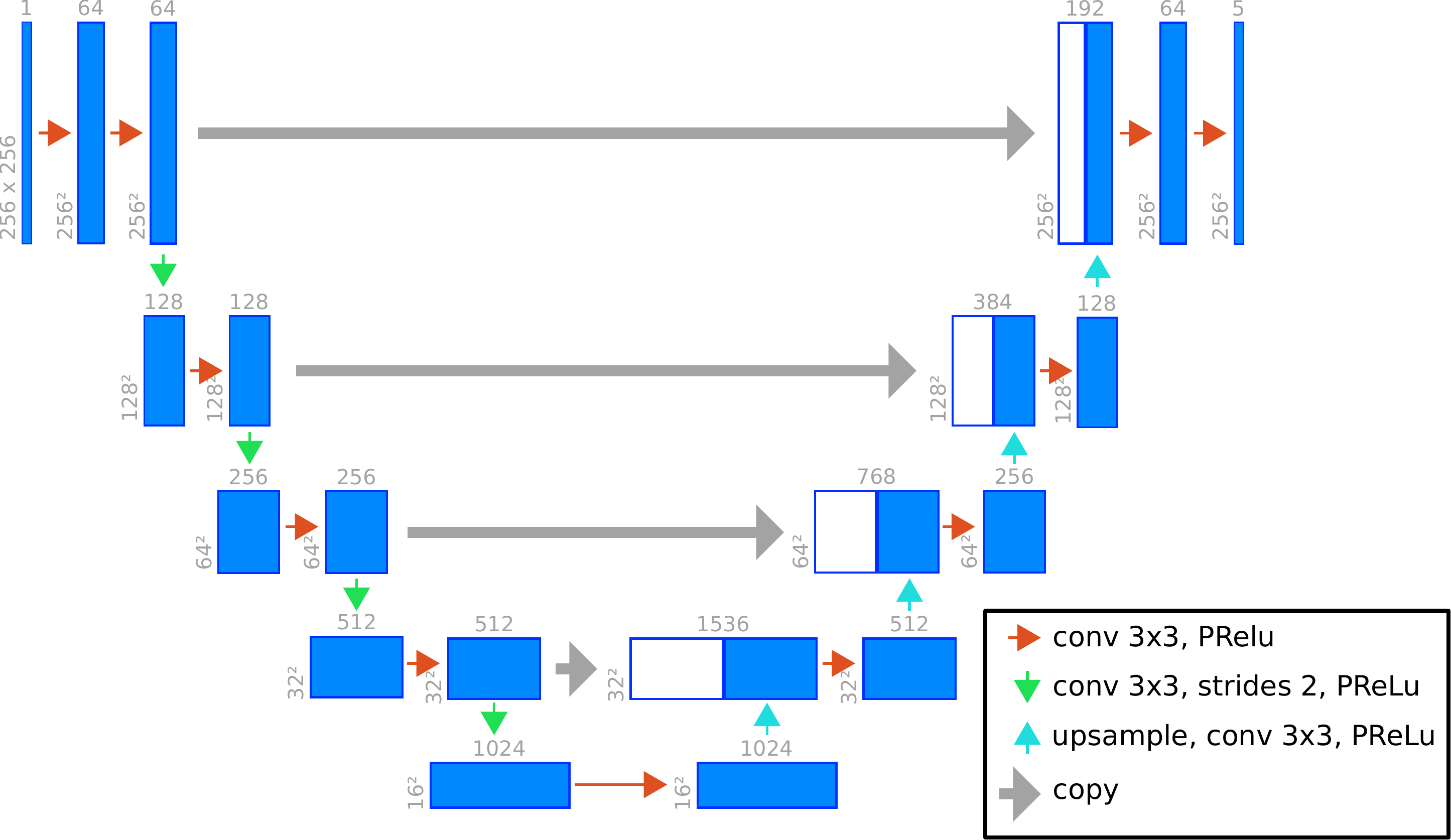}
    \caption{Topology of $\nu$-net, the artifical neural network of this study. The architecture is derived from U-Net. Each blue box corresponds to a multi-channel feature map. Size and number of channels are denoted at the upper and left of each box. Different operations are symbolized by color coded arrows.}
    \label{fig:topology}
\end{figure*}

\subsubsection{\gls{RVSC} Data Set}
The \gls{RVSC} was held at the MICCAI 2012 conference. For this challenge a data set consisting of 16 training and 32 validation 2D short axis cine \gls{MRI} stacks were acquired. Contour data is only provided for the training set, the validation contours are withheld by the authors in order to ensure an independent validation.

For this study we utilize the training set for validation, as contour information is only available for the training set. However, it has to be stressed, that no image or segmentation data of the training set has been used in the training process of the neural network. All 16 training cases have been segmented at the end-systolic and end-diastolic phase, yielding a total of 32 2D image stacks. Throughout the rest of this paper we refer to the actual training set as the validation set of this study.

Image acquisition has been performed on a 1.5~T scanner (Symphony Tim, Siemens  Medical Systems, Erlangen, Germany). Retrospectively gated balanced \gls{ssfp} cine \gls{MRI} sequences were performed for analysis with repeated breath-holds of 10-15~s. A total of 8-12 2D short axis cine planes were acquired from the base to the apex of the ventricles. The temporal resolution of the cine images is 20 images per cardiac cycle. All images have been zoomed and cropped to a 256~$\times$~216 or 216~$\times$~256 resolution, leaving the \gls{LV} visible.

\subsection{Network Topology}
The topology of $\nu$-net (/nju:n$\varepsilon$t/), the neural network of this study, is depicted in Figure~\ref{fig:topology}. It is derived from the U-Net architecture \cite{ronneberger_u-net_2015} and has been implemented in TensorFlow~\cite{tensorflow_2015}. The input layer has been resized to 256~$\times$~256 neurons with a consecutive downsampling of the subsequent layers using a 3~$\times$~3 convolution with a 2~$\times$~2 striding in contrast to a max-pooling operation of the original U-Net architecture. The padding has been changed from valid to same, resulting in an output layer of the same size as the input layer. All activation functions have been changed from a \gls{relu} to a \gls{prelu} as proposed by He et al.~\cite{he_delving_2015}.

\subsection{Network Training}
\label{nn training}
The neural network was trained by minimizing binary cross-entropy as objective function. Backpropagation was used to compute the gradients of the cross-entropy loss. The model was initialized with random values sampled from a uniform distribution without scaling variance (uniform scaling) as proposed by Glorot and Bengio~\cite{glorot_understanding_2010}. Adaptive Moment Estimation (Adam)~\cite{kingma_adam_2014} was chosen as stochastic optimization method. The initial learning rate of 10$^{-3}$ was gradually reduced down to 10$^{-6}$ during training.

The training set is assembled from 3,519 2D images (2,894 Hannover Medical School and 625 Data Science Bowl Cardiac Challenge images). Data augmentation has been applied to artificially inflate the training set as described in Section~\ref{data augmentation}. One complete training run of $\nu$-net takes about 24 to 36 hours.

\subsection{Data Augmentation}
\label{data augmentation}

\begin{figure}[t!]
    \centering
    \includegraphics[width=\linewidth]{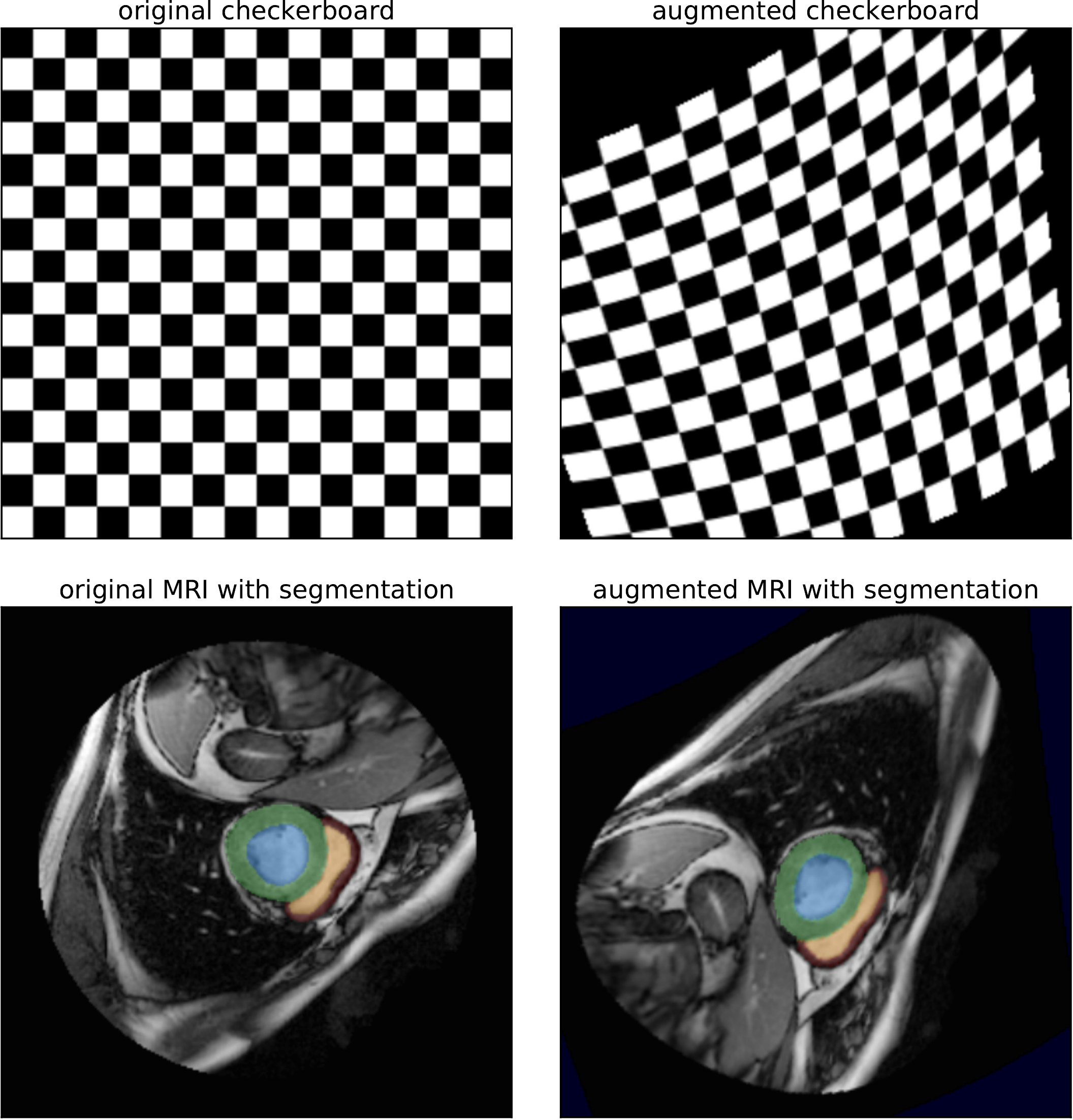}
    \caption{An example for image augmentation on the \gls{LVSC} (SC-HF-I-1) data set. The upper two panels depict a checkerboard pattern on a $256 \times 256$ domain that has been rotated, sheared, scaled, translated, randomly flipped along the axes and locally deformed ($\epsilon = 0.2$). The lower to panels apply the same augmentation to an \gls{MRI} scan with an exemplary segmentation. The augmented version is also used during training in order to artificially inflate the number of samples.}
    \label{fig:image_augmentation}
\end{figure}

Data augmentation is often used to artificially inflate the training data set~\cite{simard_best_2003, ciresan_high-performance_2011, ciregan_multi-column_2012}. This technique generates similar images and their corresponding segmentations from already existing data by applying local spatial transformations to them. The key idea behind this approach is that a slightly deformed heart should be identified by the neural network in a similar manner. Due to the image augmentation in the training phase, the algorithm provides good segmentation results regardless of the orientation, scale and parity of the input image. Translational equivariance is inherited from the convolutional network architecture, i.e. a transformed heart is mapped pixel-wise on the corresponding transformed segmentation in the spatial domain.

In order to fully utilize the computational resources of modern workstations equipped with multicore \glspl{CPU} and \gls{CUDA}-enabled \glspl{GPU}, we have developed an auxiliary library that facilitates parallel image augmentation on the \gls{CPU} while training of the neural network is delegated to the \gls{GPU}. The time needed for image augmentation can completely be hidden since concurrent augmentation of a batch of input images can be accomplished faster than the actual training step. Hence, augmentation and training are performed efficiently in an interleaved manner: the next batch of images is augmented on the \gls{CPU} while the current batch is still being trained on the \gls{GPU}.

The library is implemented in the C++ programming language and the \gls{OpenMP} extension. It features convenient bindings for the Python programming language which seamlessly interact with the TensorFlow framework. Moreover, we can apply highly non-linear, and computationally expensive local deformations to the input data as well as the ground truth segmentations, due to the aforementioned efficient interleaving. Besides traditional global transformations from the affine group $\mathop{\mathrm{Aff}}(\mathbb R, n) = \mathop{\mathrm{GL}}(\mathbb R, n) \ltimes \mathbb R^n$ such as scaling, shearing, rotation, mirroring, and translations, we allow for the pixel-wise deformation of the spatial pixel domain 
\begin{equation}
\begin{pmatrix}i \\ j \end{pmatrix} 
             \mapsto \begin{pmatrix} i+f^{(+)}(i, j) \\ j+f^{(-)}(i,j) \end{pmatrix} \quad ,
\end{equation}
where $f^{(\pm)}(i,j)$ are second degree multivariate polynomials in the pixel coordinates $i$ and $j$
\begin{alignat}{100}
f^{(\pm)}(i,j) :&= a_i^{(\pm)}\cdot i &&+ a_j^{(\pm)} \cdot j + a_{ij}^{(\pm)} \cdot ij \nonumber\\ 
             &+ a_{ii}^{(\pm)} \cdot i^2 &&+ a_{jj}^{(\pm)}  \cdot j^2 
\end{alignat}
and the coefficients $a^{(\pm)}$ are sampled from a uniform distribution over the closed interval $[-\epsilon, +\epsilon]$. The hyper-parameter $\epsilon \geq 0$ controls the amount of deformation. The special case $\epsilon = 0$ refers to no deformation. Fractional indices are mapped via bilinear interpolation. Figure~\ref{fig:image_augmentation} shows an exemplary augmentation of an \gls{MRI}.


\section{Results}

\begin{figure*}[t]
    \centering
    \includegraphics[width=\linewidth]{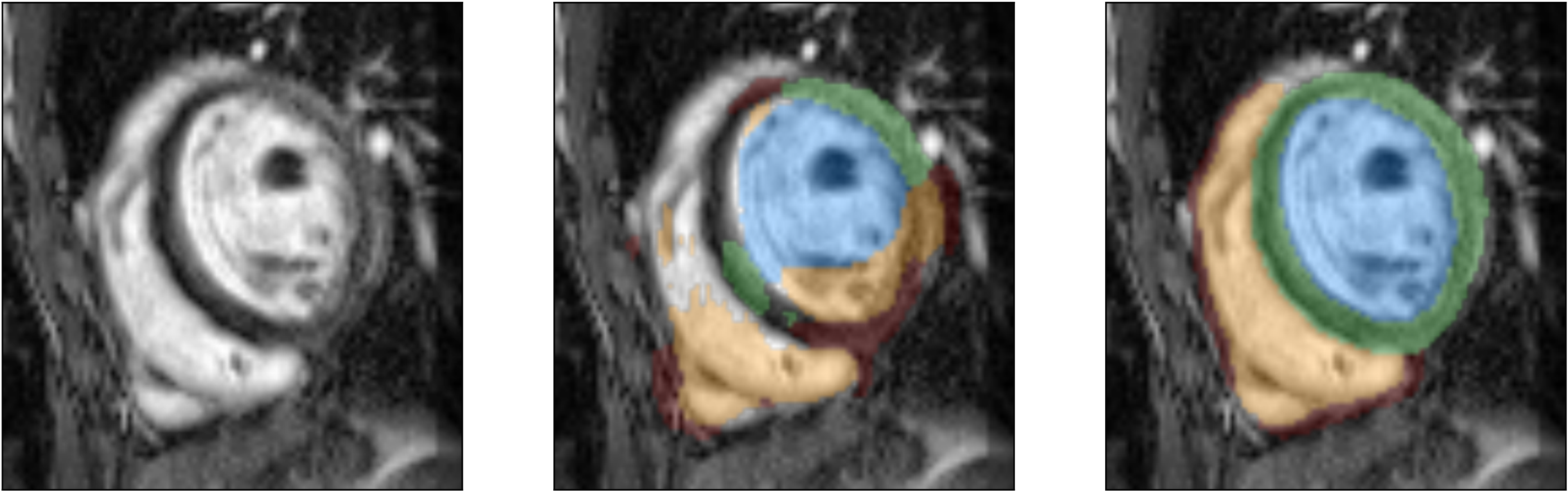}
    \caption{Depiction of the resulting prediction of an overfitted model. For this illustration $\nu$-net was trained solely on the \gls{mhh} data set. Inference was performed on an \gls{kaggle} image. From left to right, this figure shows the \gls{MRI}, the predicted segmentation with the overfitted network and the predicted segmentation with the final $\nu$-net, trained as described in Section \ref{nn training}.}
    \label{fig:overfit}
\end{figure*}

This study explores the possibility of creating a general purpose cardiac image segmentation model, capable of reliably producing high quality segmentations, independent of aspects such as different image acquisition techniques, and diverse \gls{MRI} protocols. For this purpose, the model was trained on a proprietary data set (\gls{mhh}) as well as a small subsample of the \gls{kaggle} training set. The goal was to learn the specific concept of cardiac segmentation from the highly standardized \gls{mhh} data set as well as abstracting a more general notion for different image morphologies from the heterogeneous \gls{kaggle} data set and, in turn, prevent overfitting on the characteristics of a specific data set, as depicted in Figure~\ref{fig:overfit}. The resulting intraclass correlation coefficients for all evaluated data sets are listed in Table~\ref{tab:icc}.

\begin{figure}[p]
    \centering
    \includegraphics[width=\linewidth]{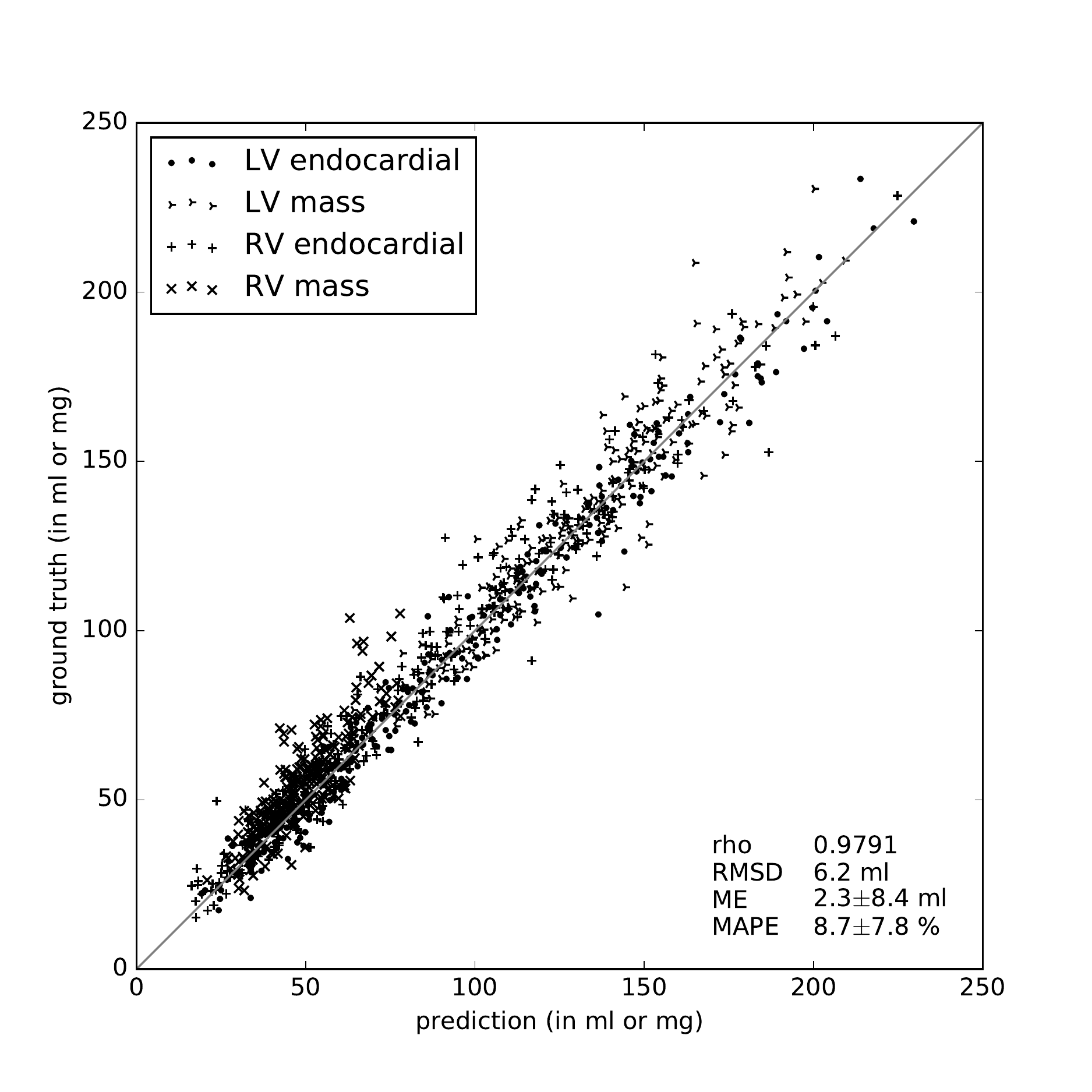}
    \caption{Correlation plot of the \gls{RV} and \gls{LV} endocardial volumes and VM of the Hannover Medical School Data Set. Spearman's rho (rho), \acrfull{RMSD}, \acrfull{ME} and \acrfull{MAPE} have been calculated overall.}
    \label{fig:mhh correlation}
\end{figure}

\begin{figure*}
    \centering
    \includegraphics[width=.75\linewidth]{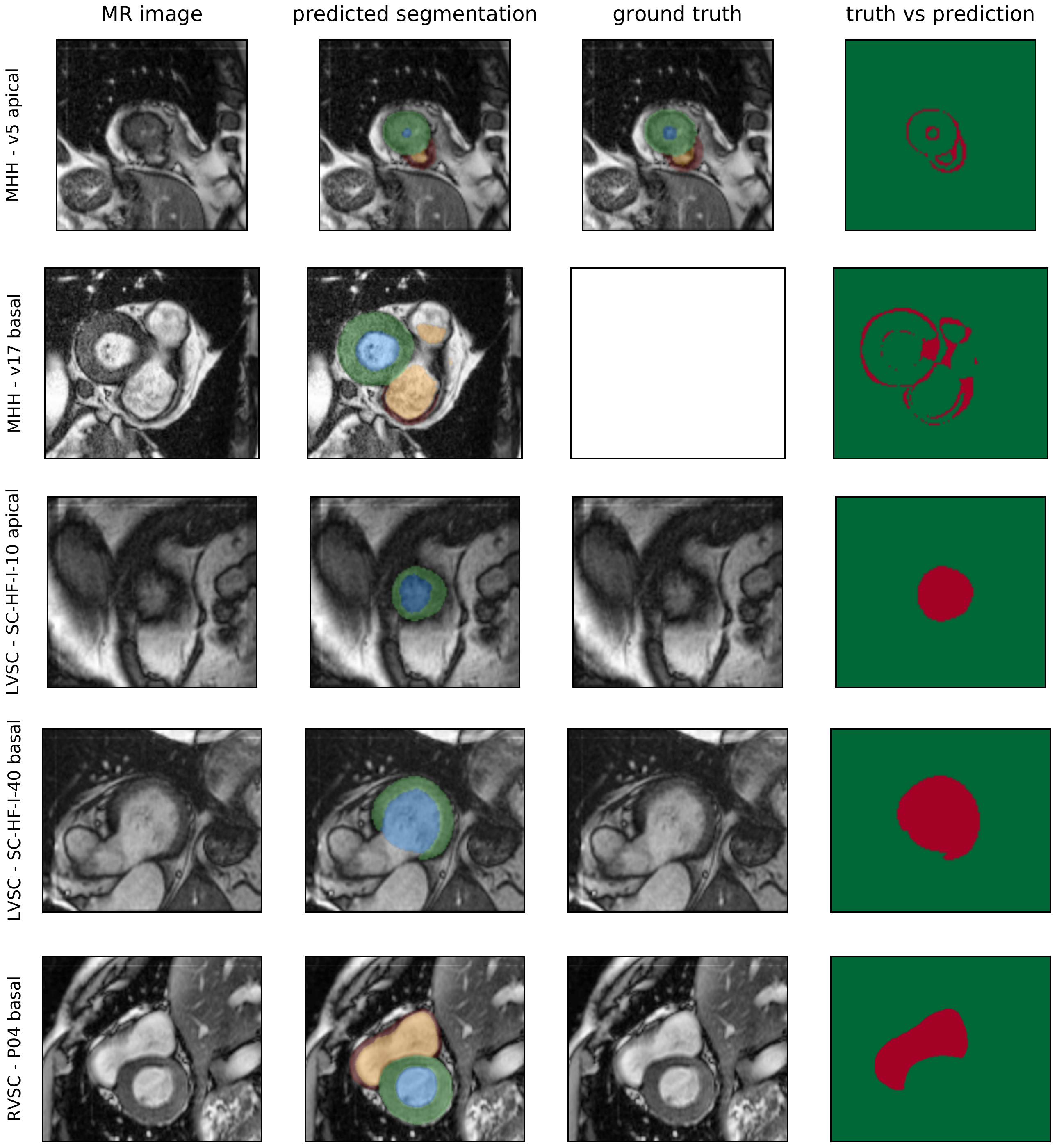}
    \caption{A selection of images of the \acrfull{mhh}, \acrfull{LVSC}, and the \acrfull{RVSC} data sets. From the left to the right column, the images depict the \gls{MRI}, the predicted segmentation, the ground truth segmentation and the difference between ground truth and prediction, where the error is denoted in red. Each case has a legend left of the image with "data set - case id and region". The upper two rows show good agreement between the predicted segmentation by $\nu$-net and the ground truth at the apex and the base. After an extensive search in the $\nu$-net \gls{mhh} results, no major disagreement between the manual and automated segmentation for the cardiac contours could be found. The lower three rows show cases at the apex and base of disagreement between predicted segmentation and ground truth. Retrospectively, one could argue that the $\nu$-net segmentation provides a more accurate delineation of the epi- and endocardium, compared to the ground truth.}
    \label{fig:bad agreement}
\end{figure*}

The agreement of the predicted segmentations with the ground truth is high for the \gls{mhh} data set with a Spearman's rho of 0.98 for an overall of \mbox{$4\cdot 309=1,236$} segmented volumes of the left and right endocardium as well as \gls{VM} as depicted in Figure \ref{fig:mhh correlation}. This is supported by a \gls{dice} of about 90~\% as depicted in Table~\ref{tab:comparison} as well as an high \gls{ICC} of 92-99 \% for the \gls{ESV} and \gls{EDV} of both ventricles. A lower \gls{dice} of 78~\% is obtained for the right \gls{VM}. Further results, such as \gls{dice}, overlap, and accuracy are listed in Table~\ref{tab:mhh results}. A selection of images of the \gls{mhh}, \gls{LVSC}, and \gls{RVSC} data sets is illustrated in Figure~\ref{fig:bad agreement}.

\begin{table}[p]
\tiny
\center
\begin{tabular}{llccc}
\toprule
 & method & V & DSC (epi) & DSC (endo) \\
\midrule
\multicolumn{5}{l}{Hannover Medical School} \\ 
 & proposed & LV & 95$\pm$2 \% & 92$\pm$4 \% \\ 
 & proposed & RV & 90$\pm$4 \% & 88$\pm$6 \% \\ 

\addlinespace
\midrule
\multicolumn{5}{l}{MICCAI 2009 LV Segmentation Challenge} \\ 
 & proposed (ad-hoc) & LV & 93$\pm$3 \% & 84$\pm$7 \% \\ 
 & proposed (with retraining) & LV & 95$\pm$3 \% & 94$\pm$3 \% \\ 
 & \citet{ngo_combining_2017} & LV & 93$\pm$2 \% & 88$\pm$3 \% \\  
 & \citet{avendi_combined_2016} & LV & n/a & 94$\pm$2 \% \\ 
 & \citet{poudel_recurrent_2016} & LV & n/a & 90$\pm$4 \% \\ 
\addlinespace
 & \citet{tran_fully_2016} & LV & 96$\pm$1 \% & 92$\pm$3 \% \\ 
 & \citet{queiros_fast_2014} & LV & 94$\pm$2 \% & 90$\pm$5 \% \\ 
 & \citet{ngo_fully_2014} & LV & n/a & 88$\pm$3 \% \\ 
 & \citet{hu_hybrid_2013} & LV & 94$\pm$2 \% & 89$\pm$3 \% \\ 
 & \citet{liu_automatic_2012} & LV & 94$\pm$2 \% & 88$\pm$3 \% \\ 
\addlinespace
 & \citet{huang_image-based_2011} & LV & 93$\pm$2 \% & 89$\pm$4 \% \\
 
\addlinespace
\midrule
\multicolumn{5}{l}{Right Ventricular Segmentation Challenge} \\ 
 & proposed (ad-hoc) & RV & 86$\pm$6 \% & 85$\pm$7 \% \\ 
 & \citet{avendi_fully_2016} & RV & n/a & 81$\pm$21 \%  \\
 & \citet{tran_fully_2016} & RV & 86$\pm$11 \% & 84$\pm$21 \% \\ 
 & \citet{zuluaga_multi-atlas_2013} & RV & 80$\pm$22 \% & 76$\pm$25 \% \\ 
 & \citet{wang2012simple} & RV & 63$\pm$35 \% & 59$\pm$34 \% \\ 
\addlinespace
 & \citet{ou_multi-atlas_2012} & RV & 63$\pm$27 \% & 58$\pm$31 \% \\
\bottomrule
\end{tabular}
\caption{
{\bf Comparison of the \gls{RV} and \gls{LV} (V) segmentation performance of the epi- and endocardium (endo) and ventricular mass (VM) based on the \acrfull{dice}.} All values are denoted as mean$\pm$std.}
\label{tab:comparison}
\end{table}

\begin{figure}[p]
    \center
    \includegraphics[width=.85\linewidth]{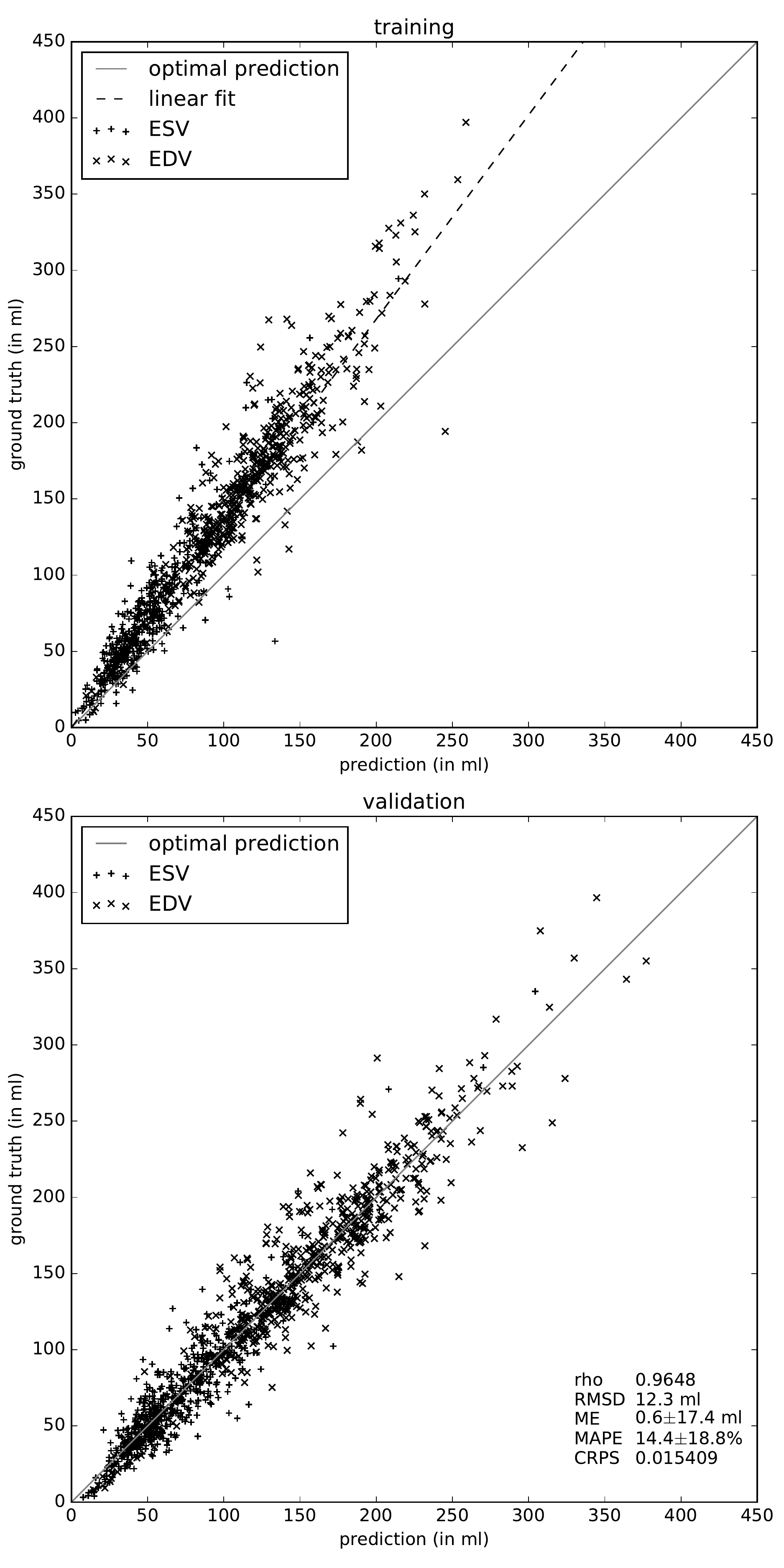}
    \caption{Correlation plot of the \gls{kaggle} ground truth versus the prediction. A small subset of the Data Science Bowl training images were manually segmented and used in the training process. Then the resulting classifier was used to predict the left ventricular volumes of the training set. The results were used to calculate the systematic error via linear regression. Then the classifier was used to segment the validation data set. The results were adjusted with the conversion factor determined by the linear regression. The adjusted results are plotted in the lower subfigure. Spearman's rho (rho), \acrfull{RMSD}, \acrfull{ME} and \acrfull{MAPE} have been calculated overall.}
    \label{fig:kaggle correlation}
\end{figure}

The predicted segmentation can be used to directly compute the clinical parameters on \gls{mhh} data by applying the Simpson’s method~\cite{winter_evaluating_2008}. The same  approach cannot be performed on \glspl{MRI} stemming from different sources, because of different segmentation philosophies, resulting in differing clinical measurements for the same case. To compensate for this variation, a linear regression has been performed to adapt the predicted \gls{ESV} and \gls{EDV} of the \gls{kaggle} data set. This fit has been determined using standard linear regression, mapping the predicted volumes of the training set onto the ground truth scalar volumes, whilst omitting the vertical intercept. Neither the model in training, nor the linear regressor were fitted to any of the validation cases in order to eliminate the possibility of leakage. As depicted in Figure~\ref{fig:kaggle correlation} the agreement with the ground truth is high with a Spearman's rho of 0.96, an \gls{ICC} of 92/94~\% (\gls{ESV}/\gls{EDV}), and a \gls{MAPE} of 14~\%.

The performance of the proposed method would have ranked under the best 20 competitors of the Second Annual Data Science Bowl with a CRP score of 0.0154. Figure~\ref{fig:kaggle worst} depicts the weakest segmentation, based on the CRP score, for additional illustration.

The correlation of the left and right epi- and endocardial volumes of the \gls{LVSC} and \gls{RVSC} data sets is high with a Spearman's rho of 0.95 as depicted in Figures~\ref{fig:sunnybrook_lv} and \ref{fig:sunnybrook_rv} as well as \gls{ICC} values of 96/97~\% (\gls{ESV}/\gls{EDV}) for the left and 92/96~\% for the right ventricle. A lower rho of 0.83 is expected for the right epicardium. \gls{dice} for a selection of relevant studies are presented in Table~\ref{tab:comparison}. \gls{RMSD}, \gls{ME} and \gls{MAPE} were calculated after adjusting the resulting volumes with a linear regression, fitted on the validation-set. This was necessary, as the validation sets are too small to perform a reasonable split. A random sample of images and the corrosponding segmentation from the \gls{LVSC} and \gls{RVSC} data sets is illustrated in Figures~\ref{fig:lvsc random samples} and \ref{fig:rvsc random samples}.

\begin{table*}
\center
\begin{tabular}{lccccc|cc}
\toprule
 & \multicolumn{2}{c}{MHH} & DSBCC & LVSC & RVSC & \multicolumn{2}{c}{HUMAN} \\
\cmidrule{2-3} \cmidrule{7-8} & RV & LV & LV & LV & RV & RV & LV\\
\midrule
EF   & \textbf{0.960} & \textbf{0.983} & 0.794 & 0.945 & \textbf{0.867} & 0.800 & 0.953 \\
EDV  & \textbf{0.958} & 0.985 & 0.935 & 0.966 & \textbf{0.924} & 0.892 & 0.987 \\
ESV  & \textbf{0.920} & 0.953 & 0.918 & 0.962 & \textbf{0.958} & 0.917 & 0.992 \\
SV   & \textbf{0.923} & \textbf{0.978} & \textbf{0.898} & \textbf{0.907} & \textbf{0.841} & 0.814 & 0.867 \\
VM & \textbf{0.832} & \textbf{0.948} & & \textbf{0.941} & \textbf{0.825} & 0.54 & 0.848 \\
\bottomrule
\end{tabular}
\caption{Comparison of the \acrfull{ICC} for the clinical parameters EF~(\%), EDV~(ml/m$^2$), ESV~(ml/m$^2$), SV~(ml/m$^2$), and VM~(g/m$^2$) between the ground truth and the fully automatically measured volumes. Human inter-observer \gls{ICC} (denoted as human) has been determined by Caudron et al.~\cite{caudron_cardiac_2012}. Coefficients, surpassing human performance, have been highlighted. \gls{LVSC} and \gls{RVSC} are reported as ad-hoc performance.}
\label{tab:icc}
\end{table*}

\section{Discussion}
There are many philosophies on how to perform image segmentation of cardiac \glspl{MRI}. These philosophies have major influence on the resulting biventricular mass and function parameters. For example, the inclusion or exclusion of trabeculations and papillary muscles affect the left and right ventricular mass as well as the endocardial volume~\cite{sievers_impact_2004, winter_evaluating_2008, vogel-claussen_left_2006}. No convention has been universally accepted for analyzing trabeculation and papillary muscle mass~\cite{schulz-menger_standardized_2013}.

Typically, these conventions are enforced on an institute basis. This poses a major challenge for the automated image segmentation with deep neural networks, because of the resulting difference of the clinical measurements. In order to learn the specific segmentation characteristics of a site, a specifically tailored model would have to be trained. This renders the whole process impractical.

However, most of the time, image segmentation is only means to the end of determining the clinical parameters by measuring the ventricular volumes. These parameters include the \acrfull{VM}, \acrfull{ESV} as well as \acrfull{EDV}, which yield the \acrfull{EF} and \acrfull{SV}. These established clinical parameters are of great importance in early detection of cardiac illnesses as well as treatment monitoring.

The hypothesis of this study is that the clinical parameters can be reliably measured by adapting a pre-trained neural network to a new environment and applying one of the most basic statistical models, a linear regression. This result is unexpected since differences in segmentation guidelines are usually of local nature and do not necessarily need to exhibit a linear dependency on the final measurement. In order to substantiate this assumption, $\nu$-net was trained on the extensive \gls{mhh} data set. $\nu$-net demonstrates state-of-the-art performance, as depicted in the results Figure~\ref{fig:mhh correlation}. Furthermore, $\nu$-net was trained on a small, hand-picked subset of the \gls{kaggle} training set. This was done with the idea of transfer learning in mind, in order to convey an understanding of different image acquisition methods, and varying image morphologies to the neural network.

$\nu$-net was benchmarked against the \gls{kaggle} validation set. As depicted in Figure~\ref{fig:kaggle correlation}, it was possible to adapt the results of the classifier for the specific data set by employing a linear regression, fitted on the training set.

The performance of the proposed method would have ranked under the best 20 competitors of the Second Annual Data Science Bowl. A strong correlation between predicted and ground truth volumes was observed with a Spearman's rho of 0.965 and a \gls{MAPE} of 14.4~\%. Nevertheless, the final predicted volumes of the proposed solution could be fine-tuned using high-level machine learning regressors as performed by the winning solution. This study, however, explicitly omits sophisticated post-processing and use of meta data (such as gender, age, height, and scanner geometry) in order to avoid over-fitting to a specific experimental setting. This could reasonably impair the neural network's generalization potential.

The hypothesis is further substantiated by the results of the \gls{LVSC} and \gls{RVSC} data sets. As depicted in Figures~\ref{fig:sunnybrook_lv} and \ref{fig:sunnybrook_rv} the segmentation results demonstrate a strong correlation with the manual segmentation exhibiting a Spearman's rho of approximately 0.95. Furthermore, the corresponding \glspl{ICC} imply human level performance in determining clinical parameters. $\nu$-net surpasses human level in predicting parameters for the right ventricle. Note that volume prediction was performed without training on a single image of these data sets.

$\nu$-net demonstrates a state-of-the-art ad-hoc segmentation performance in terms of \gls{dice} for the epi- and endocardium of the \gls{RVSC} and the epicardium of the \gls{LVSC} data set compared to \citet{tran_fully_2016} and \citet{avendi_fully_2016} as well as other studies~\cite{avendi_combined_2016, ngo_combining_2017, poudel_recurrent_2016, queiros_fast_2014, hu_hybrid_2013, liu_automatic_2012, huang_image-based_2011, zuluaga_multi-atlas_2013, wang2012simple, ou_multi-atlas_2012}. This is remarkable, as $\nu$-net was not trained on any images of the aforementioned data sets. The slightly weaker ad-hoc \gls{dice} of 84~\% for the endocardium of the \gls{LVSC} could reflect different segmentation philosophies compared to the original MHH data set, resulting in a systematic error as shown in Figure~\ref{fig:sunnybrook_lv}. This hypothesis is supported by corresponding high ad-hoc \gls{ICC} values for the \gls{ESV}, \gls{EDV}, \gls{EF}, and \gls{SV}. Furthermore, additional retraining results in state-of-the-art performance for the endocardial \gls{dice}.

Regarding the MHH data set, $\nu$-net achieves comparable or higher agreement with the ground truth than two human observers agree on average in measuring biventricular mass and function parameters~\cite{caudron_cardiac_2012}. Furthermore, $\nu$-net accomplishes comparable to human performance on the \gls{LVSC} and \gls{RVSC} data sets. $\nu$-net outperforms a human by a wide margin especially at the task of gauging the right ventricular endocardial volume and ventricular mass. A slightly lower \gls{ICC} score of the left endocardial volumes on the \gls{kaggle} data set is most likely due to a multi-center and multi-observer setting, resulting in a inherent heterogeneity in the data set. In order to improve the performance, the results would have to be evaluated for each observer independently.

One limitation of this study is the small size of openly available data sets. The \gls{LVSC} and \gls{RVSC} contain 61 cases with freely accessible contours. Furthermore, the aforementioned data sets include the segmentation of the \gls{LV} or \gls{RV} exclusively. Additionally, training and validating on a single center data set bears the risk of overfitting as demonstrated in Figure~\ref{fig:overfit}. Therefore, a large, fully labeled, multi-center, multi-reader data set would be advantageous. This data set could be used to train and evaluate future models.


\section{Conclusion}
This study demonstrates the reliability of automatically determining clinical cardiac parameters such as \gls{ESV}, \gls{EDV}, \gls{EF}, \gls{SV} and \gls{VM} on 4 data sets. Especially in the \gls{RV} the neuronal network outperformed the human cardiac expert in the presented study, which likely enables more reliable \gls{RV} mass and function measurements for improved clinical treatment monitoring in the future. Furthermore, it is demonstrated that the aforementioned parameters, resulting of an image segmentation by a pre-trained neural network, can be adjusted by performing a linear regression. This effectively eliminates the associated costs of introducing a neural network for determining clinical cardiac parameters in a new setting.

\subsection{Perspectives}
\noindent \textbf{COMPETENCY IN MEDICAL KNOWLEDGE:}
Deep learning can reliably determine high quality fully-automated cardiac segmentation for precise determination of clinically well established biventricular mass and function parameters in a multi center setting.

\noindent \textbf{TRANSLATIONAL OUTLOOK:} 
The presented neuronal network is ready to be used in large scale, multi center, multi reader data sets for cost- and time efficient analysis of cardiac mass and function parameters.

\section{Acknowledgments}
The authors thank all parties involved in the data acquisition process, especially Frank Schr\"oder and Lars K\"ahler.

\onecolumn

\clearpage
\appendix

\section{Figures and Tables}

\begin{figure}
    \centering
    \includegraphics[width=.9\linewidth]{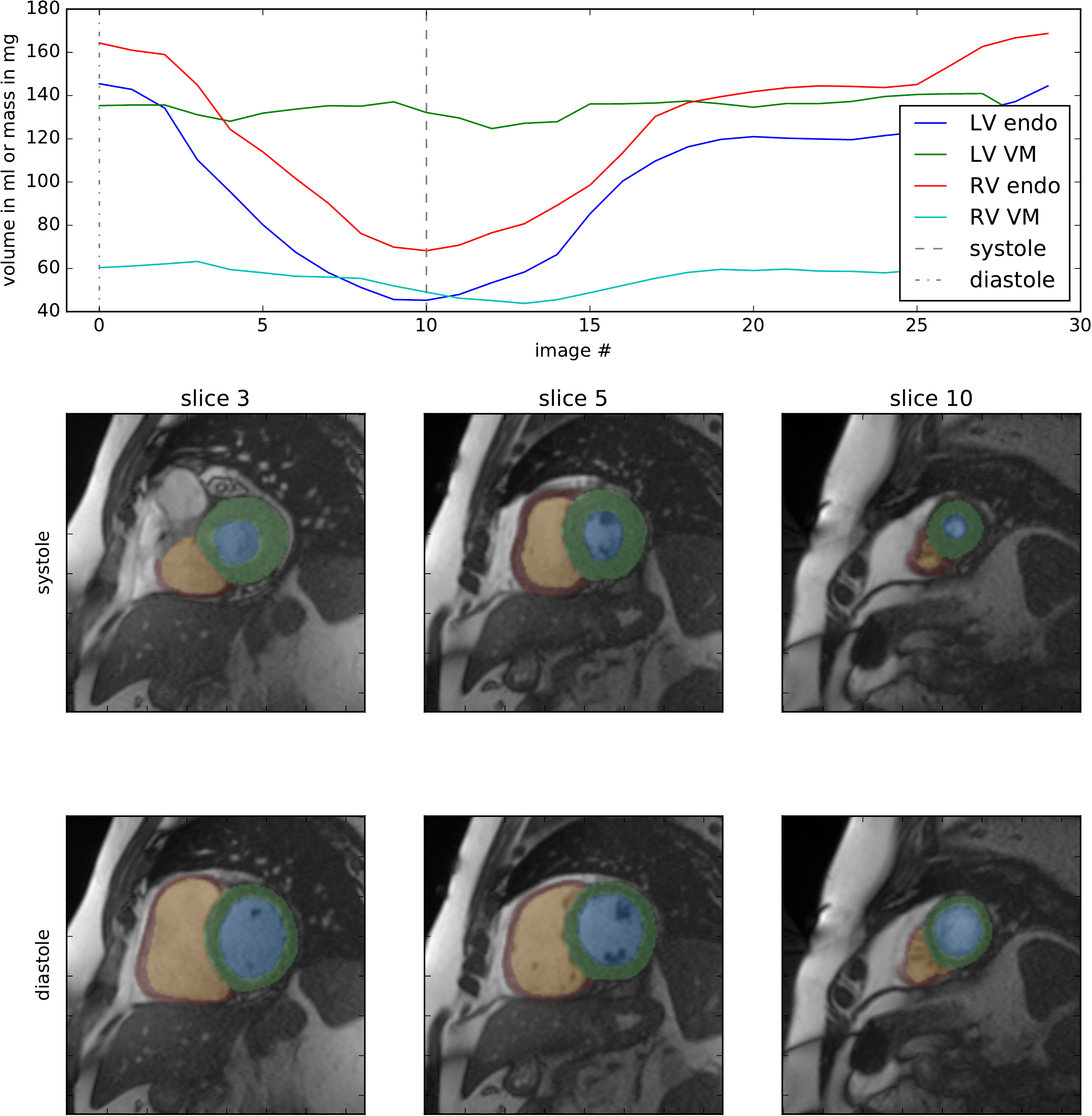}
    \caption{Depiction of the automated image segmentation results for the case with the worst systolic (0.095) and diastolic (0.129) CRP score in the \gls{kaggle} data set. The upper plot illustrates the measured volumes for the left and right epi- and endocardium against time. The image number (\#) for end-systole and end-diastole were determined automatically by $\nu$-net. End-systole and end-diastole were identified automatically. The ground truth for this sample is 127 ml systolic and 291.4 ml diastolic. However, the segmentation suggest an end-systolic volume of 64.5 ml and an end-diastolic volume of 198 ml.}
    \label{fig:kaggle worst}
\end{figure}

\begin{figure}
    \centering
    \includegraphics[width=.7\linewidth]{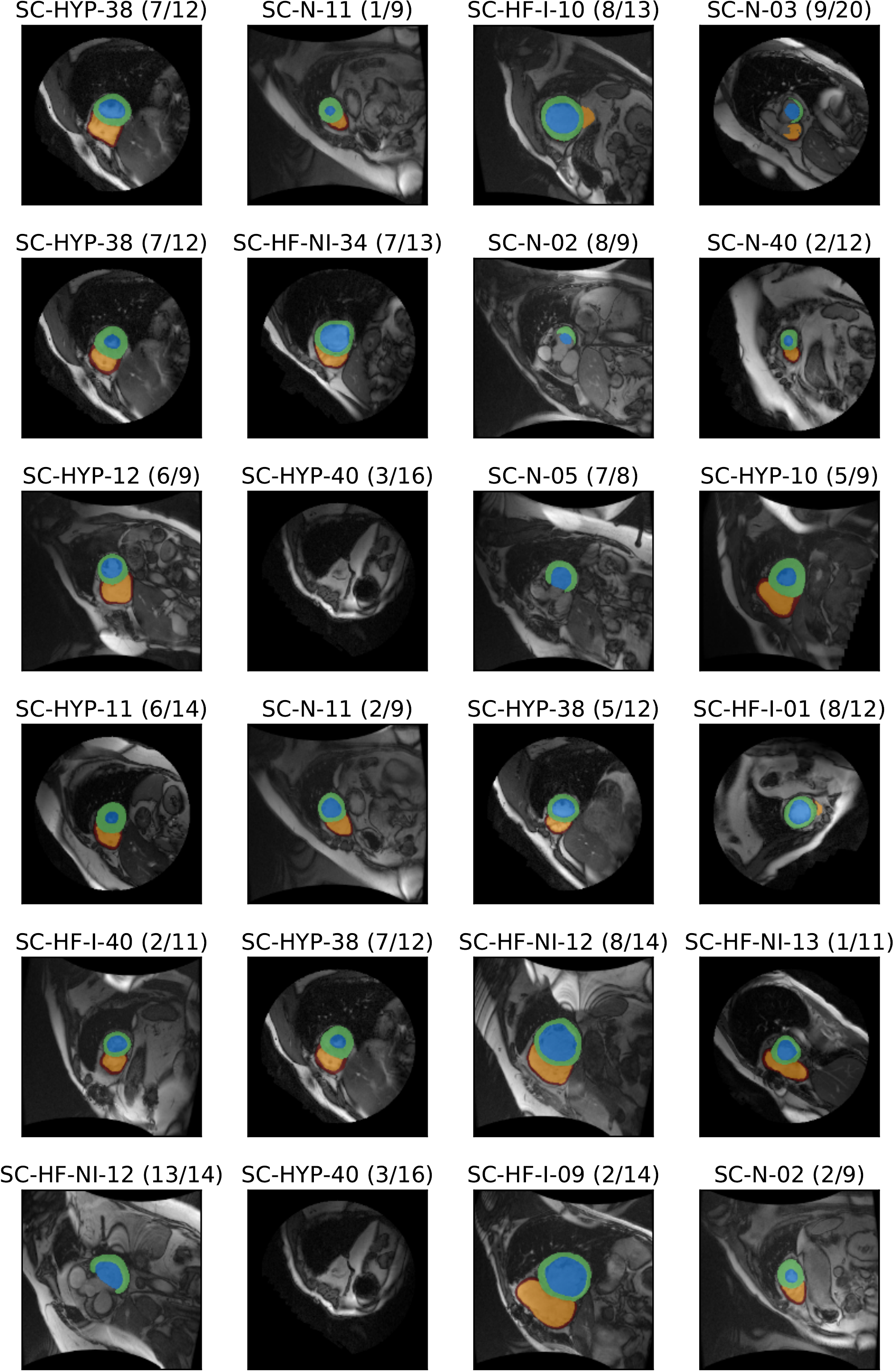}
    \caption{24 randomly chosen samples of the predicted segmentation of $\nu$-net for the \gls{LVSC} data set. The title of each sub-figure denotes the case identifier and the randomly selected image slice in brackets.}
    \label{fig:lvsc random samples}
\end{figure}

\begin{figure}
    \centering
    \includegraphics[width=.7\linewidth]{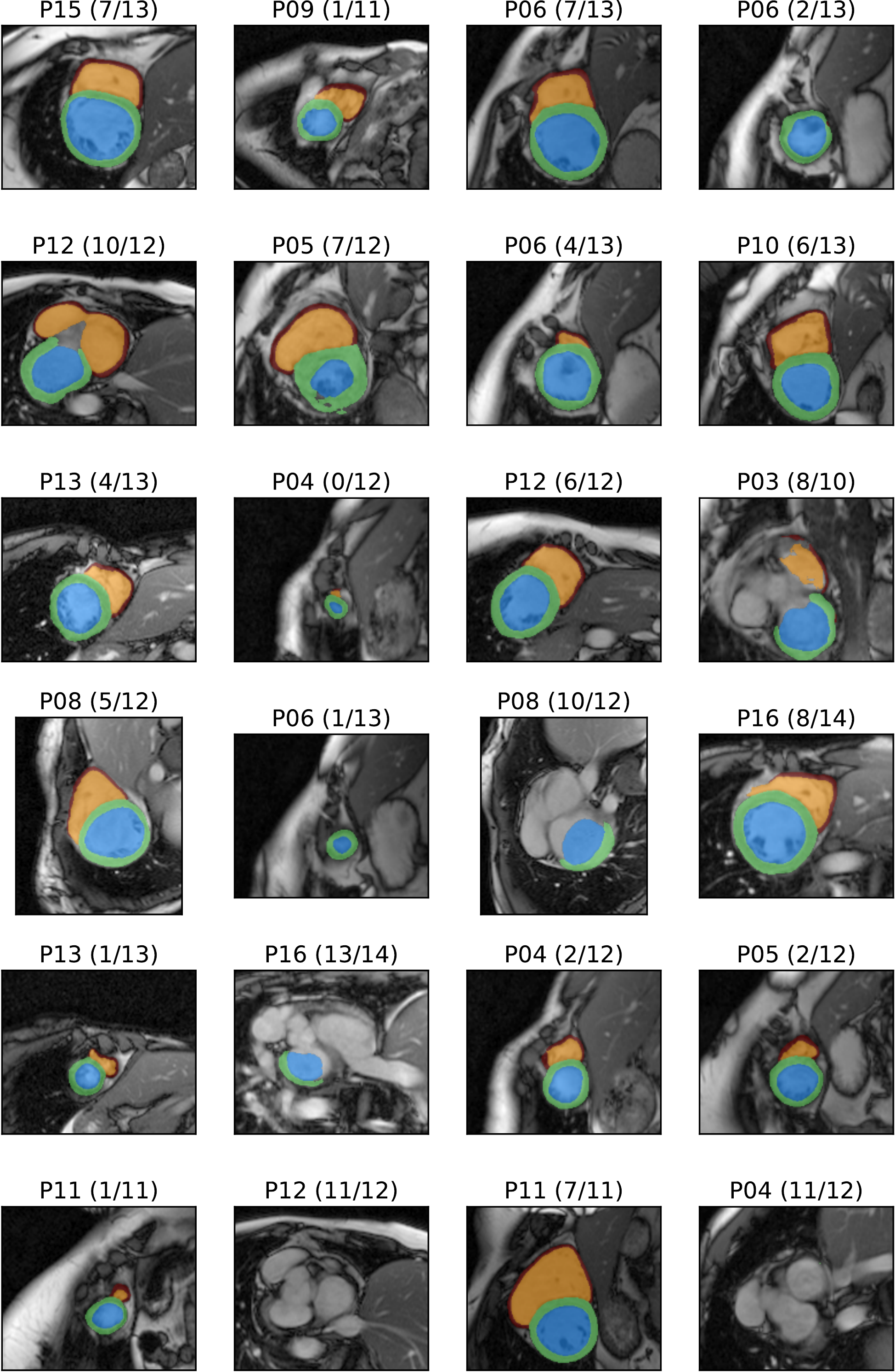}
    \caption{24 randomly chosen samples of the predicted segmentation of $\nu$-net for the \gls{RVSC} data set. The title of each sub-figure denotes the case identifier and the randomly selected image slice in brackets.}
    \label{fig:rvsc random samples}
\end{figure}

\begin{figure*}
    \centering
    \includegraphics[width=0.8\linewidth]{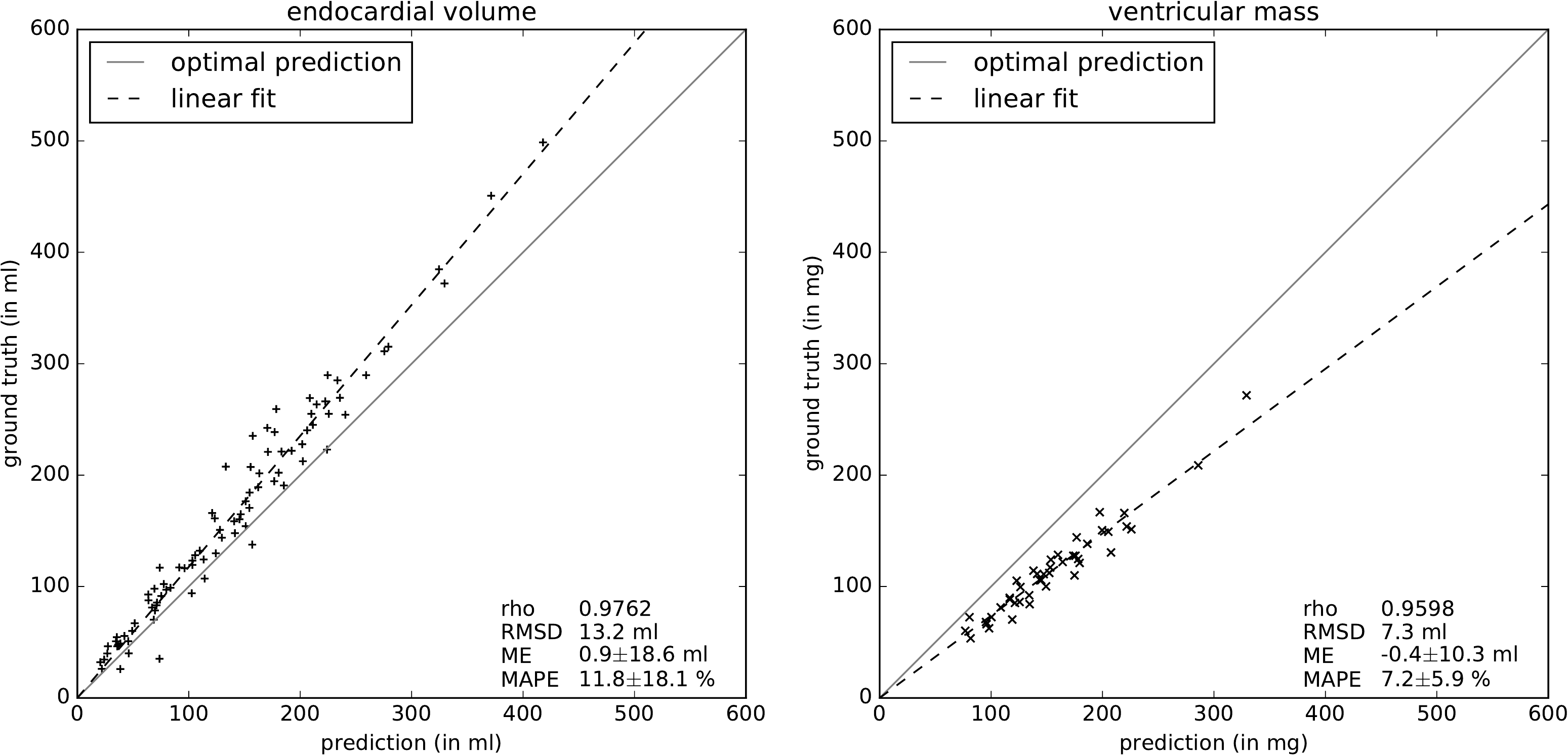}
    \caption{Correlation graph of the ad-hoc performance of $\nu$-net for the \acrfull{LVSC} data set, depicting the LV endocardial volume and ventricular mass. Spearman's rho (rho), \acrfull{RMSD}, \acrfull{ME} and \acrfull{MAPE} have been calculated after applying the linear fit.}
    \label{fig:sunnybrook_lv}
\end{figure*}

\begin{figure*}
    \centering
    \includegraphics[width=0.8\linewidth]{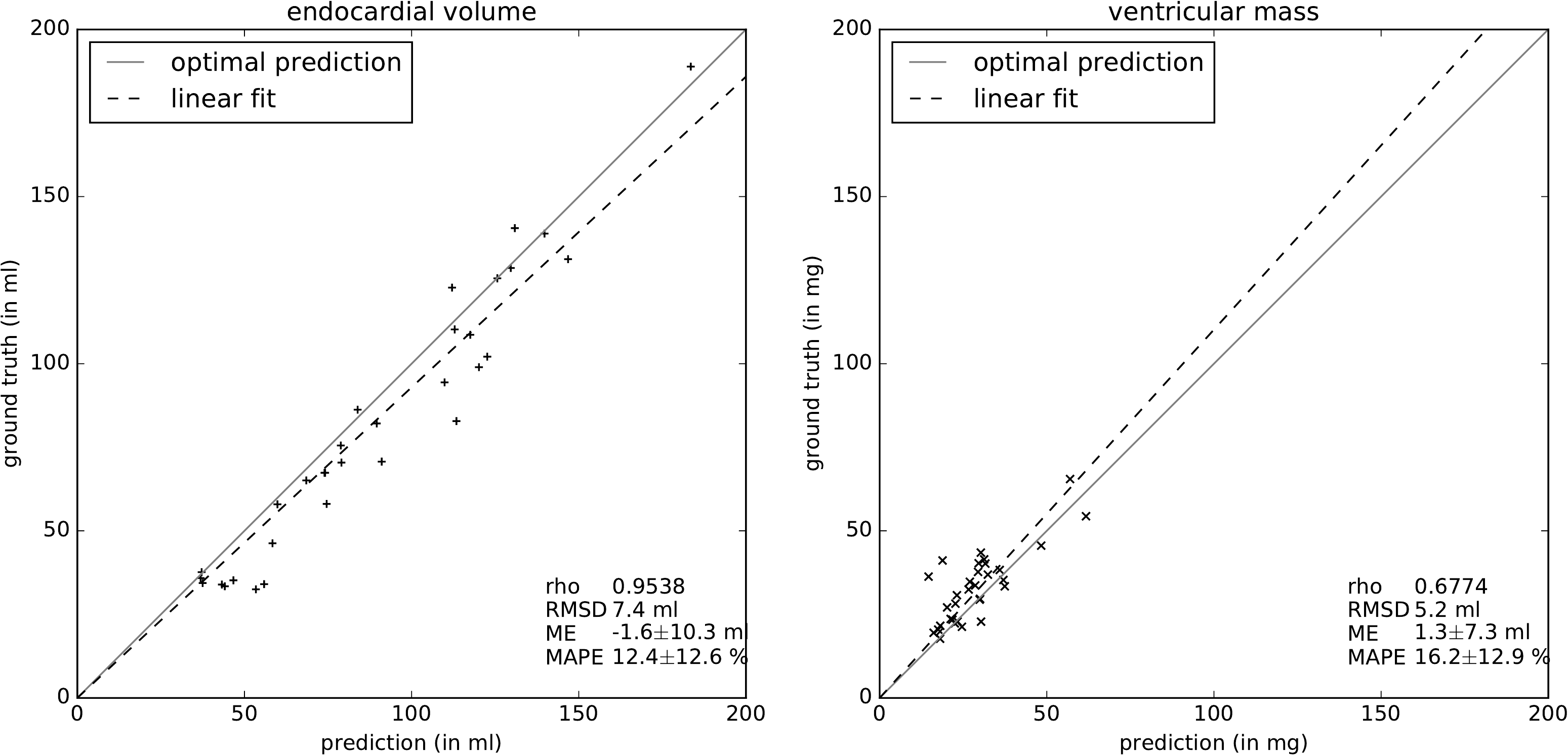}
    \caption{Correlation graph of the ad-hoc performance of $\nu$-net for the \acrfull{RVSC} data set, depicting the LV endocardial volume and ventricular mass. Spearman's rho (rho), \acrfull{RMSD}, \acrfull{ME} and \acrfull{MAPE} have been calculated after applying the linear fit.}
    \label{fig:sunnybrook_rv}
\end{figure*}

\begin{table}[h]
\begin{tabular}{lllllll}
\toprule
label & accuracy & specificity & precision & recall & dice & overlap \\
\midrule
LV endocardial & 99.9$\pm$0.0 \% & 99.9$\pm$0.0 \% & 91.1$\pm$6.6 \% & 92.3$\pm$5.1 \% & 91.5$\pm$4.3 \% & 84.6$\pm$6.9 \% \\
LV mass & 99.8$\pm$0.1 \% & 99.9$\pm$0.0 \% & 89.6$\pm$4.3 \% & 88.3$\pm$4.0 \% & 88.8$\pm$2.9 \% & 80.1$\pm$4.6 \% \\
LV epicardial & 99.8$\pm$0.1 \% & 99.9$\pm$0.1 \% & 94.8$\pm$3.4 \% & 94.3$\pm$2.7 \% & 94.5$\pm$2.0 \% & 89.6$\pm$3.4 \% \\
\midrule 
RV endocardial & 99.9$\pm$0.1 \% & 99.9$\pm$0.0 \% & 89.4$\pm$6.2 \% & 86.7$\pm$7.9 \% & 87.7$\pm$5.5 \% & 78.5$\pm$8.1 \% \\
RV mass & 99.8$\pm$0.1 \% & 99.9$\pm$0.0 \% & 80.4$\pm$8.2 \% & 72.4$\pm$7.1 \% & 75.8$\pm$5.9 \% & 61.4$\pm$7.5 \% \\
RV epicardial & 99.8$\pm$0.1 \% & 99.9$\pm$0.0 \% & 92.9$\pm$4.1 \% & 87.2$\pm$5.9 \% & 89.8$\pm$3.8 \% & 81.8$\pm$6.0 \% \\
\bottomrule
\end{tabular}
\caption{Descriptive statistical analysis of the results based on the Hannover Medical School Data-Set. All values have been computed at the end-systolic and end-diastolic phase. All values are denoted as mean$\pm$std.}
\label{tab:mhh results}
\end{table}

\clearpage

\section{Bibliography}
\printbibliography

\end{document}